\pdfoutput=1
\documentclass[11pt]{article}

\usepackage{acl}

\usepackage{times}
\usepackage{latexsym}

\usepackage[T1]{fontenc}
\usepackage[utf8]{inputenc}

\usepackage{microtype}

\usepackage{inconsolata}

\usepackage{kotex}
\usepackage{graphicx}
\usepackage{multirow}
\usepackage{pifont}
\newcommand{\cmark}{\ding{51}}%
\newcommand{\xmark}{\ding{55}}%
\usepackage{tabularray}
\usepackage{array}
\usepackage{amsmath, amssymb}

\DeclareMathSymbol{\mlq}{\mathord}{operators}{``}
\DeclareMathSymbol{\mrq}{\mathord}{operators}{`'}
\usepackage{subcaption}
\usepackage{booktabs}
\usepackage{colortbl}
\usepackage{soul}
\usepackage{xcolor}
\usepackage{soulutf8}
\definecolor{skyblue}{rgb}{0.5, 0.8, 1.0}
\definecolor{pink}{rgb}{1.0, 0.75, 0.79}
\usepackage{tcolorbox}
\newtcolorbox{important_blue}{
    colframe=skyblue!50,%
    colback=skyblue!50,%
    left=1pt, right=1pt,%
    top=0.5pt, bottom=0.5pt,%
    boxsep=0pt,%
    hbox,
    before=\vspace{0em},
    after=\vspace{0em}
}
\newtcolorbox{important_red}{
    colframe=pink!50,%
    colback=pink!50,%
    left=1pt, right=1pt,%
    top=0.5pt, bottom=0.5pt,%
    boxsep=0pt,%
    hbox,
    before=\vspace{0em},
    after=\vspace{0em}
}

\usepackage{setspace}
\title{MP2D: An Automated Topic Shift Dialogue Generation Framework Leveraging Knowledge Graphs}

\author{Yerin Hwang\textsuperscript{1} \hspace{1.3cm} Yongil Kim\textsuperscript{2}\hspace{1.3cm} Yunah Jang\textsuperscript{2}\hspace{1cm} \\ \textbf{Jeesoo Bang}\textsuperscript{3}  \hspace{1cm} {\bf Hyunkyung Bae\textsuperscript{3}} \hspace{1cm}  {\bf Kyomin Jung\textsuperscript{1,2,4$\dagger$}} \\
  $^{1}$IPAI, Seoul National University
  $^{2}$Dept. of ECE, Seoul National University\\
  $^{3}$LG AI Research
  $^{4}$SNU-LG AI Research Center\\
  \texttt{\{dpfls589, miles94, vn2209, kjung\}@snu.ac.kr}\\
  \texttt{\{jeesoo.bang, hkbae\}@lgresearch.ai}
  }

\begin{document} 
\maketitle
\begin{abstract}
Despite advancements in on-topic dialogue systems, effectively managing topic shifts within dialogues remains a persistent challenge, largely attributed to the limited availability of training datasets.  
To address this issue, we propose Multi-Passage to Dialogue (MP2D), a data generation framework that automatically creates conversational question-answering datasets with natural topic transitions.
By leveraging the relationships between entities in a knowledge graph, MP2D maps the flow of topics within a dialogue, effectively mirroring the dynamics of human conversation.
It retrieves relevant passages corresponding to the topics and transforms them into dialogues through the passage-to-dialogue method. 
Through quantitative and qualitative experiments, we demonstrate MP2D’s efficacy in generating dialogue with natural topic shifts. 
Furthermore, this study introduces a novel benchmark for topic shift dialogues, TS-WikiDialog. Utilizing the dataset, we demonstrate that even Large Language Models (LLMs) struggle to handle topic shifts in dialogue effectively, and we showcase the performance improvements of models trained on datasets generated by MP2D across diverse topic shift dialogue tasks.

\end{abstract}

\section{Introduction}

\begin{figure}[t]
\centering
\includegraphics[width= 0.87\columnwidth]{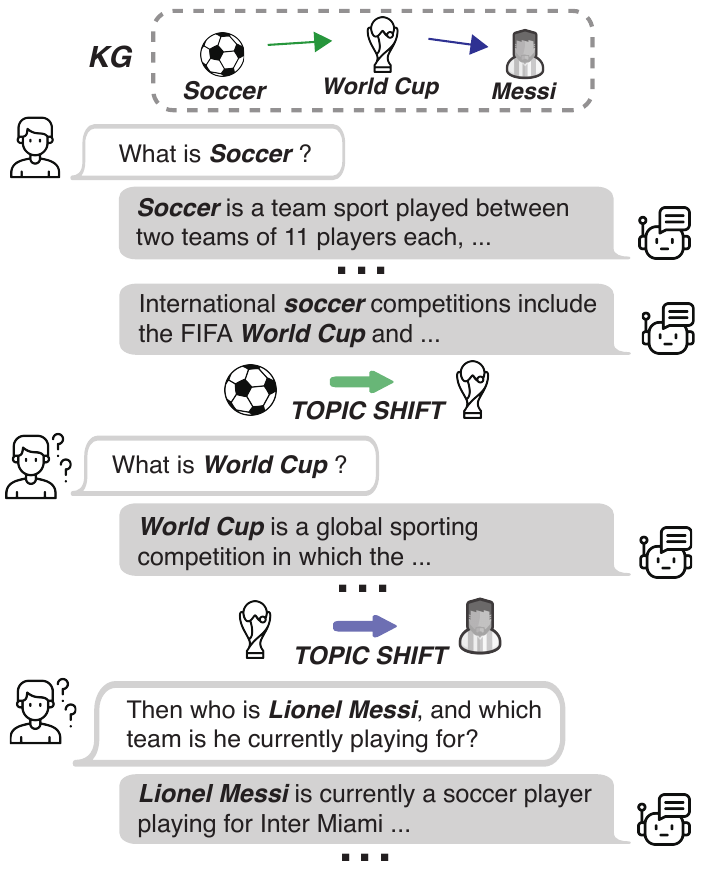} 
\caption{An example of a topic shift dialogue. The MP2D framework utilizes paths in a Knowledge Graph (KG) to extract entities and facilitates natural topic transitions based on the relations between these entities.}
\label{figure1}
\vspace{-5mm}
\end{figure}
%Dialog systems, aiming to respond to queries or provide relevant information to users, have gained significant attention from both academia and industry. These systems hold potential applications across various fields, including virtual assistants, conversational experiences, and interactive chatbots. However, while many existing studies have advanced the field of conversational systems within on-topic dialog scenarios, there continues to be a significant challenge regarding dialogs involving topic shifts.

Dialogue systems~\cite{chen2017survey, ni2023recent}, designed to respond to inquiries or provide relevant information, have gained considerable attention in academia and industry. These systems show promise in various applications, including virtual assistants, customer service, and chatbots~\cite{ king2023future,zeng2024automated,liu2024dragon}.
Nonetheless, while numerous studies have advanced the field of conversational systems within on-topic dialogue scenarios, significant challenges persist concerning dialogues encompassing topic shifts~\cite{holtzman2019curious, zhang2019grounded, xu2021topic}.
 
%Unlike human conversation, which can naturally transition between topics and continue the discussion seamlessly, existing dialog system models often struggle to gauge the timing for topic shifts or to execute these shifts fluently.This issue is compounded by the fact that creating all existing topic shift dialog datasets involves a time-consuming and expensive human annotation process.

%In contrast to the fluidity inherent in human conversation, which seamlessly transitions between topics and sustains discourse, current dialog system models often grapple with discerning the appropriate timing for topic shifts or executing these shifts seamlessly. 
Unlike human conversation, which can naturally flow between topics and continue the discussion seamlessly, current dialogue systems often struggle to determine the appropriate timing for topic shifts or to execute these shifts fluently~\cite{xie2021tiage}.
A major challenge in these tasks lies in the scarcity of data. This issue is compounded by the fact that creating all existing topic shift dialogue datasets~\cite{xu2021topic, sevegnani2021otters} involves a laborious and costly human annotation process.
%A primary hurdle in topic shift dialogues is the scarcity of data, compounded by the laborious and costly human annotation process required for creating existing topic shift datasets.

%In response to this challenge, we propose a novel framework, Multi-Passage to Dialog (MP2D), designed for the automatic generation of topic shift conversational question-answering datasets(ConvQA). The framework employs the passage-to-dialog method for ConvQA dataset generation. However, unlike existing methods that generate dialogs from a single passage, this framework integrates multiple passages to produce dialogs with natural topic shifts. To facilitate such shifts, akin to those occurring in real-world conversations, this approach utilizes a knowledge graph to identify paths connecting various entities and their relations. By retrieving passages using entities in the path as queries, the passages and the relation sentences between entities form a multi-passage structure. This structure is then segmented into sentences, directly used as answers, and suitable questions are generated by a question generator to complete dialogs that incorporate natural topic shifts in the conversation.

To address this challenge, we propose a novel framework, Multi-Passage to Dialogue (MP2D), specifically crafted for the automatic generation of Conversational Question-Answering (ConvQA) data featuring topic shifts.
The framework leverages the Passage-to-Dialogue (P2D) method~\cite{dai2022dialog, hwang2023dialogizer} in ConvQA dataset generation, which entails generating relevant questions by using the sentences within a passage as answers.
Unlike existing methods that construct dialogues from a single passage, our proposed framework integrates multiple passages to create dialogues with topic shifts.
To emulate the dynamic nature of real-world conversations, we employ a knowledge graph to identify paths connecting various entities and their relations, as shown in Figure~\ref{figure1}. 
By retrieving passages using entities in the path as queries, the passages and the relation sentences between entities form a multi-passage structure.
%Retrieving passages using entities in the path as queries, we construct a multi-passage structure comprising passages and relation sentences between entities. 
%This structure is then segmented into sentences, serving directly as answers. Suitable questions are generated by a question generator to complete dialogues, ensuring the incorporation of natural topic shifts in the conversation. 
%This structure is then segmented into sentences, which directly serve as answers, and suitable questions are generated by a question generator to complete dialogues with natural topic transitions.
This structure is then segmented into sentences, which directly serve as answers, and a question generator generates suitable questions to complete dialogues with natural topic transitions.
%The framework adopts Large Language Models (LLMs)~\cite{brown2020language} as question generators and generates suitable questions for the given answers to complete dialogs, ensuring the incorporation of natural topic shifts in the conversation.

%We conduct experiments applying various question generators for multi-passage to dialog transformations, utilizing various reference-free dialog metrics for automatic evaluation. The results reveal that while existing passage to dialog methods are effective in converting single passages into dialogues, Large Language Models (LLMs) demonstrate superior performance in seamlessly transitioning multi-passage inputs into topic shift dialogues. Furthermore, we validate the high quality of datasets generated through the MP2D framework. Through qualitative assessments conducted using human evaluation and GPT-4 evaluation, we found that approximately 91\% of the dialogs were deemed excellent in their handling of topic shifts. 

%We conduct experiments utilizing various question generators for multi-passage to dialog transformations, employing diverse reference-free dialog metrics for automatic evaluation. 
We conduct experiments using various question generators to transform multi-passage into dialogue, employing diverse reference-free dialogue metrics for automatic evaluation.
%The results indicate that while existing P2D methods effectively convert single passages into dialogs, Large Language Models (LLMs)\cite{brown2020language} demonstrate superior performance in seamlessly transitioning multi-passage inputs into dialogs with topic shifts. 
%The results indicate that Large Language Models (LLMs)\cite{brown2020language} demonstrate superior performance in seamlessly transitioning multi-passage inputs into dialogs with topic shifts. 
%Furthermore, we validate the high quality of datasets generated through the MP2D framework. 
%The results indicate that Large Language Models (LLMs)~\cite{brown2020language} are effective in generating dialogs with natural topic shifts, and that MP2D-generated datasets possess high quality.
%The results indicate that MP2D-generated topic shift dialog datasets possess high quality.
%The results indicate that the LLM as a question generator module is effective, and the MP2D-generated topic shift dialog datasets possess high quality.
The results indicate that the generated topic shift dialogue datasets demonstrate high quality when utilizing Large Language Models (LLMs)~\cite{brown2020language} as a question generator within MP2D.
Moreover, through qualitative assessments conducted via human and GPT-4 evaluation~\cite{liu2023gpteval, wang2023chatgpt}, we found that approximately  91\% of the generated dialogues are deemed excellent in their handling of topic shifts.

%Additionally, we introduce a novel topic shift dialog benchmark, TS-Wikidialog, to address the weaknesses of LLMs in handling topic shift dialog tasks and to broaden the application of datasets generated via MP2D. The benchmark is generated by paraphrasing the MP2D-generated dialog by LLMs and human annotators. In experiments utilizing the TS-Wikidialog, we initially verify that various LLMs struggle to effectively address problems related to topic segmentation, topic shift detection, and topic-shift conversational question answering. Furthermore, the T5 model trained on a training dataset generated through MP2D, demonstrates significantly better performance compared to baseline LLMs, thereby validating the MP2D framework as an effective dataset creation framework for addressing topic shift problems. Additionally, we demonstrate that using a fine-tuned model, trained on the MP2D-generated training dataset for the topic segmentation task, can enhance the performance of LLM responses in topic shift turns.

%Furthermore, we introduce a novel topic shift dialog benchmark for evaluating LLMs in handling topic shift dialog tasks and extending MP2D-generated dataset’s applicability. 
Furthermore, we introduce a novel topic shift dialogue benchmark, TS-WikiDialog, to evaluate LLMs in handling topic shift dialogue tasks and to broaden the application of datasets generated via MP2D. 
The benchmark is constructed by paraphrasing the MP2D-generated dialogue by LLMs and human annotators.
%Furthermore, the T5 model~\cite{raffel2020exploring}, trained on a training dataset generated through MP2D, exhibits significantly better performance in both topic segmentation and topic shift detection tasks compared to baseline LLMs, thereby validating the MP2D framework as an effective dataset creation approach for tackling topic shift challenges.
In experiments utilizing the TS-WikiDialog, we initially verify that various LLMs struggle to effectively address problems related to topic segmentation~\cite{purver2011topic}, topic shift detection~\cite{holz2010towards}, and topic shift ConvQA~\cite{xie2021tiage}.
%Furthermore, the T5 model~\cite{raffel2020exploring}, trained on MP2D-generated datasets, exhibits significantly better performance in topic segmentation and topic shift detection tasks compared to baseline LLMs, thereby validating the MP2D framework as an effective dataset creation approach for tackling topic shift challenges.
Furthermore, the T5-base model~\cite{raffel2020exploring}, trained on the MP2D-generated dataset, exhibits significantly better performance in topic segmentation and topic shift detection tasks compared to baseline LLMs. This validates the MP2D framework as an effective dataset-creation approach for tackling topic shift challenges.
%Additionally, we demonstrate that fine-tuning models on MP2D-generated datasets for the topic segmentation task can enhance the performance of LLM responses in topic shift turns. 
Additionally, we demonstrate that utilizing finetuned models on MP2D-generated datasets for the topic shift detection task can enhance the performance of LLM responses in ConvQA, especially in topic shift turns.

\begin{figure*}[t]
\centering
\includegraphics[width=0.82\textwidth]{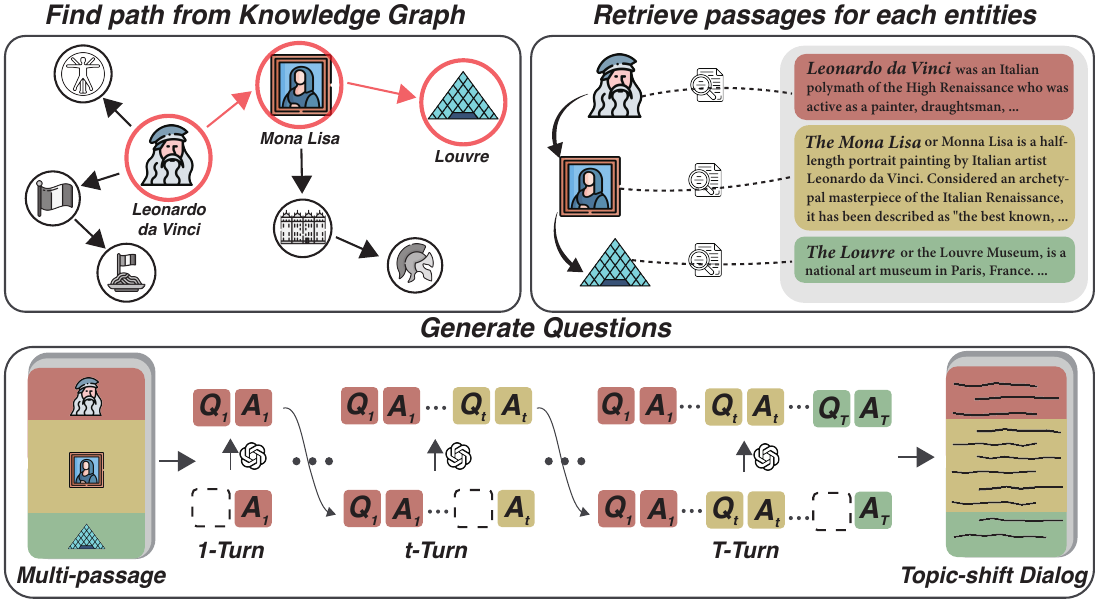} 
\caption{An overview of the MP2D framework. In the knowledge graph, paths are identified and passages are retrieved for entities within those paths. Then, the retrieved passages and their relations become the \textit{"answers"}, and a LLM generates \textit{"questions"} corresponding to each answer to create dialogues.} 

\label{figure2}
\vspace{-3.5mm}
\end{figure*}
\section{Related Works}
\subsection{Topic Shift Dialogue Systems}

%Topic shift dialog denotes instances within multi-turn dialogues where the conversation's focus changes mid-discussion. Such topic switching is a natural aspect of human interaction. According to \citet{soni2021empirical}, a topic shift occurs on average every 12 turns. Particularly in ConvQA, which aims to deliver information to the user, the occurrence of users changing topics is more commonly observed\cite{spink2002multitasking} and poses a major challenge\cite{zaib2023learning}. This typically happens because users pose follow-up questions to explore new curiosities that arise from previous answers. 
%For example, in a conversation initially focused on 'soccer,' if the 'World Cup' emerges in one of the responses, the discussion may then shift to the 'World Cup,' eventually leading the conversation to the captain of the 2022 World Cup-winning country, 'Lionel Messi.'

Topic shift dialogue refers to instances within multi-turn dialogues where the focus of the conversation changes mid-discussion~\cite{garcia1997analysis}.
Such transitions are a natural aspect of human interaction. 
%According to \citet{soni2021empirical}, a topic shift occurs, on average, every 12 turns.
According to \citet{soni2021empirical}, a change in topic happens every 12 conversational turns.
Particularly in ConvQA, which aims to provide information to users~\cite{zaib2022conversational}, the phenomenon of users changing topics is more commonly observed~\cite{spink2002multitasking} and presents a significant challenge~\cite{zaib2023learning}.
%This often arises because users pose follow-up questions to explore new curiosities that stem from previous answers. 
Topic transitions in ConvQA often occur because users pose follow-up questions to explore new curiosities that stem from previous answers.
%For example, in a conversation initially centered around ”soccer,” if the “World Cup” emerges in one of the responses, the discussion may then shift to the “World Cup,” eventually leading to topics such as the captain of the 2022 World Cup-winning country, “Lionel Messi.”

%Dialog systems proficient in handling topic shifts must effectively detect such shifts\cite{galley2003discourse, somasundaran2020two} and maintain the quality of their responses thereafter\cite{yang2022take}; however, current conversational systems struggle with this task\cite{holtzman2019curious, xie2021tiage}. Although the data scarcity problem is a major challenge for topic shift dialog or ConvQA tasks, all existing frameworks for generating topic shift dialog datasets involve a process of human annotation. These methods have limitations; 1. they are unstable due to the subjective criteria on the scope of topics, which can vary among annotators, 2. and they are labor-intensive and time-consuming. This study proposes the first framework capable of automatically generating a conversational question-answering dataset with natural topic transitions based on a knowledge graph.

Dialogue systems proficient in handling topic shifts must effectively detect such transitions ~\cite{galley2003discourse, somasundaran2020two} and maintain the quality of their responses thereafter~\cite{wang2021topicrefine}.
However, current conversational systems struggle with this task~\cite{holtzman2019curious, xie2021tiage}.
Despite the data scarcity problem being a major challenge for topic shift dialogue or ConvQA tasks, all existing frameworks for generating topic shift dialogue datasets~\cite{xie2021tiage, yang2022take} involve a process of human annotation.
%These methods have limitations: 1. They are prone to instability due to the subjective criteria on the scope of topics, which can vary among annotators~\cite{galley2003discourse}, and 2. they are labor-intensive and time-consuming.
These methods have limitations: 1. They are prone to instability due to the subjective criteria on the scope of topics~\cite{galley2003discourse}, and 2. they are labor-intensive and time-consuming.
To the best of our knowledge, this study is the first to propose a framework capable of automatically generating a ConvQA dataset with natural topic transitions.

\subsection{Passage to Dialogue (P2D)}

%Recently, passage to dialog frameworks have been proposed to address the data scarcity issue in ConvQA by automatically generating ConvQA datasets from passages. These frameworks segment passages into sentence units to serve as "answers'' and employ question generators trained on task-specific objectives to produce corresponding "questions" for each "answer." [dialog inpainting paper] first proposes a passage to dialog framework, employing a question generator trained through a dialog reconstruction task that involves masking at the utterance level and subsequently learning to fill these gaps. In addition, [dialogizer paper] introduces additional tasks capable of learning sentence-level alignment, thereby proposing a passage to dialog framework that focuses on contextual relevance. 

%Recently, passage-to-dialog frameworks have emerged to address the data scarcity issue in ConvQA by autonomously generating ConvQA datasets from passages. 
%Recently, amoung various dataset generation methods proposed to address the issue of data scarcity in ConvQA, the passage to dialog (P2D) frameworks hold potential as they enable the generation of dialogs from textual sources without loss of information.

Recently, among various research efforts addressing the issue of data scarcity in ConvQA~\cite{dalton2020trec, anantha2020open, kacupaj2021paraqa, mulla2023automatic}, the Passage to Dialogue (P2D) frameworks hold potential as they enable the generation of dialogues from textual sources without loss of information.
These frameworks segment passages into sentence units to function as "answers" and employ question generators trained on task-specific objectives to generate corresponding "questions" for each "answer."
%For instance, \citet{dai2022dialog} first proposed a P2D framework, employing a question generator trained through a dialog reconstruction task.
\citet{dai2022dialog} first proposed a P2D framework, employing a question generator trained through a dialogue reconstruction task.
% that involves masking at the utterance level and subsequently learning to fill these gaps.
Additionally, \citet{hwang2023dialogizer} introduces supplementary tasks capable of learning sentence-level alignment, presenting a P2D framework that prioritizes contextual relevance. 

%These passage to dialog methods offer the advantage of leveraging specialized passages, which are abundantly available online, to create ConvQA datasets without any loss of information. However, these methods are limited to converting a single passage into a dialog, lacking the ability to control topic shifts or dialog flow in the generated dialogs. This study proposes the first automatic framework capable of transforming multi-passage content into dialogs with natural topic shifts.

%These methods leverage specialized passages, abundant online, to create ConvQA datasets without loss of information.
%These methods are capable of creating ConvQA datasets with specialized passages without any loss of information.
%These methods are capable of transforming specialized passages into dialogues without any loss of information, even without the participation of a domain expert.
%However, they are limited to converting a single passage into a dialogue, lacking the capability to control topic shifts or dialogue flow in the generated dialogues.
%This study introduces the first automatic framework capable of transforming multi-passage content into dialogues with natural topic shifts.
These methods can transform specialized passages into dialogues without any loss of information, even without the participation of a domain expert. 
However, their application is limited to converting a single passage into a dialogue, lacking the capability to manage topic shifts or control the flow of the generated dialogues. 
This study introduces the first automatic framework designed to transform multi-passage content into dialogues with natural topic shifts.
\section{Multi-Passage to Dialogue (MP2D)}
%Multi-Passage to Dialog (MP2D) is a novel framework designed to generate dialogs with topic shifts, utilizing multiple textual passages. By utilizing a knowledge graph, MP2D extracts paths to create natural linkages among topics, thereby establishing a basic dialog flow for smooth transitions. The framework retrieves passages by querying each entity within these paths, and a multi-passage structure is formed by pairing these passages with sentences that describe the relationships between entities. Subsequently, it segments the multi-passage into sentence-sized units to serve as answers. Leveraging the capability of LLMs to generate contextually appropriate questions for each answer, MP2D employs LLMs as question generators in crafting dialogs. An automatic post-processing phase follows, resulting in dialogs that seamlessly incorporate natural topic shifts. Figure 2 provides an illustrative overview of the proposed framework.

The Multi-Passage to Dialogue (MP2D) framework represents a novel approach tailored to generating dialogues with topic shifts by incorporating multiple textual passages. 
Through the utilization of a knowledge graph, MP2D extracts paths to establish natural linkages among topics, facilitating smooth transitions within the dialogue.
The framework retrieves passages by querying each entity within these paths, forming a multi-passage structure by pairing these passages with sentences that describe the relationships between entities~(\S\ref{3.1}).
Subsequently, it segments the multi-passage into sentence-sized units to serve as answers. Leveraging the capability of LLMs to generate contextually appropriate questions for each answer, MP2D employs LLMs as question generators in crafting dialogues~(\S\ref{3.2}). 
An automatic post-processing phase follows, resulting in dialogues that seamlessly transition between topics. Figure~\ref{figure2} provides an illustrative overview of the proposed framework.

\subsection{Find Path \& Retrieve Passages}
\label{3.1}
%The first step in our process is to map out a path through the knowledge graph that links various entities together. This helps in determining the subjects for our discussion. For this task, we use the KELD dataset, which is designed around an entity-driven knowledge graph. In this graph, knowledge is organized as triplets $(s,r,o)$: a subject entity $s$ (e.g., "Leonardo da Vinci"), a relation $r$ (e.g., "painted"), and an object entity $o$ (e.g., "Mona Lisa"). Each triplet is further enriched with sentences $R$ that provide a detailed narrative of the relationship between two entities, offering more context than the simple relation $r$. The MP2D framework employs this element to reconstruct the knowledge graph $K := {(e_i, R_i, o_i)}^N_{i=1}$ of $N$ factual triplets.

In our methodology’s initial phase, we navigate a knowledge graph to establish connections among diverse entities, determining discourse subjects. 
We utilize the KELM dataset~\cite{agarwal2020knowledge}, structured around an entity-driven knowledge graph. 
Within this graph, knowledge is arranged into triplets $(S, r, O)$: a subject entity $S$ (e.g., "Leonardo da Vinci"), a relation $r$ (e.g., "painted"), and an object entity $O$ (e.g., "Mona Lisa"). 
%Each triplet is augmented with sentence $R$, which furnish a comprehensive narrative regarding the relationship between two entities, imparting more nuanced context than the simple relation $r$.
%Each triplet is further enriched with sentences $R$ that provide a detailed narrative of the relationship between two entities, offering more context than the simple relation $r$.
The dataset additionally contains a relation sentence, denoted as $R$, which provides a detailed narrative of the relationship between two entities, offering more context than the simple relation $r$.
The MP2D framework leverages this component to reconstruct the knowledge graph $K := {(S_i, R_i, O_i)}^N_{i=1}$ of $N$ factual triplets.

%The framework selects a single $(s, R, o^*)$ triplet, enabling the connection of more than two entities, by identifying an $(o^*, R, o)$ where $o*$ becomes the subject. This process is conducted auto-regressively to construct the path $p$, iterating until it encounters a point where no further triplets exist for the last object to transition into a subject. Thus, the path $p$ is denoted as $\{s_1, R_1, o_1=s_2, R_2, o_2=s_3, ... ,o_n\}$, and by unifying subjects and objects as entities $e$, the path can be expressed as $p = \{e_1, R_1, e_2, ... e_n\}$.

%The framework selects a particular $(s, R, o^*)$ triplet, facilitating the connection of more than two entities, by identifying an $(o^*, R, o)$ where $o^*$ assumes the role of the subject. 
The framework randomly selects a particular $(S, R, O^*)$ triplet, facilitating the connection of more than two entities by identifying an $(O^*, R, O)$ where $O^*$ becomes the subject.
%This process is conducted auto-regressively to construct the path $p$, iterating until it reaches a point where no further triplets exist for the last object to transition into a subject. 
%Consequently, the path $p$ is represented as $\{S_1, R_1, O_1=S_2, R_2, O_2=S_3, ... ,O_n\}$, and by merging subjects and objects as entities denoted as $e$, the path is expressed as $P = \{e_1, R_1, e_2, R_2, ... , e_n\}$.
This process is conducted auto-regressively to construct the finite walk $\phi$, iterating until it reaches a point where no further triplets exist for the last object to transition into a subject. 
Consequently, the walk $\phi$ is represented as a path $\{S_1, R_1, O_1=S_2, R_2, O_2=S_3, ... ,O_n\}$, and by merging subjects and objects as entities denoted as $e$, the path is expressed as $\phi = \{e_1, R_1, e_2, R_2, ... , e_n\}$.

%For each entity $e_i$ along the identified path, when it is used as a query $q_i$, passages $p_i$ are sequentially retrieved. Each retrieved passage $p_i$ consists of $m$ sentences, represented as $p_i = \{s_1, s_2, ..., s_m\}$. To adapt these passages for processing, we truncate each to a maximum length of $k_i$ sentences, resulting in a truncated passage $p_i^\dagger = \{s_1, s_2, ..., s_{k_i}\}$, where $k_i = \min(\text{length}(p_i), \text{random}(3,6))$. To facilitate a natural connection within the dialog flow, sentences representing the entities' relations, denoted as $R$, are interspersed among the retrieved passages for each entity, thereby constructing a multi-passage $P = \{p_1^\dagger, R_1, p_2^\dagger,R_2, ... p_n^\dagger\}$.

%For each entity $e_i$ identified along the path, when employed as a query $q_i$, passages $p_i$ are retrieved sequentially as described in Figure~\ref{figure2} (top-right). 
For each entity $e_i$ identified along the path, using it as a query $q_i$ results in the sequential retrieval of passages $p_i$, as described in Figure~\ref{figure2} (top-right). 
Each retrieved passage $p_i$ consists of $m_i$ sentences, denoted as $p_i = \{s_1, s_2, ..., s_{m_i}\}$. 
To prepare these passages for processing, we truncate each to a maximum length of $k_i$ sentences, resulting in a truncated passage $p_i^\dagger = \{s_1, s_2, ..., s_{k_i}\}$, where $k_i = \min(m_i, \text{random}(3,6)) (k_i \ll m_i)$. 
Ultimately, alongside the existing relation sentence $R$ for a natural connection between the entities, a multi-passage \textit{MP} $= \{p_1^\dagger, R_1, p_2^\dagger, R_2, ..., p_n^\dagger\}$ is constructed.
%To establish a natural connection within the dialog flow, sentences representing the entities' relations, designated as $R$, are interspersed among the retrieved passages for each entity, thus constructing a multi-passage $P = \{p1^\dagger, R1, p2^\dagger ... pn^*\dagger\}$.

%Utilizing a knowledge graph in constructing the dialog flow within the MP2D framework presents several advantages. First and foremost, knowledge graphs are inherently designed to encapsulate the relationships between various entities, offering a structured method to comprehend the context surrounding each entity. In the MP2D framework, this facilitates the selection of topics that are not only relevant to the ongoing dialog but also interconnected in meaningful ways. This interconnectedness ensures that the dialog flows logically from one topic to the next, mirroring natural human conversations where topics shift smoothly based on underlying relationships.

The utilization of a knowledge graph in constructing the dialogue flow provides several advantages. 
Primarily, knowledge graphs inherently encapsulate relationships between various entities, offering a structured method for understanding the context surrounding each entity. 
In the MP2D framework, this facilitates the selection of topics that are not only pertinent to the ongoing dialogue but also interconnected in meaningful ways. 
This interconnectedness ensures that the dialogue flows logically from one topic to the next, mirroring natural human conversations where topics shift smoothly based on underlying relationships.

%Additionally, the use of a knowledge graph for the automatic generation of dialogs offers the distinct advantage of effortlessly creating conversations that remain current and up-to-date. As information changes over time, manually correcting dialogs can be challenging due to their inherently unstructured data. However, knowledge graphs are dynamic entities that grow and evolve over time, and substantial research is being conducted on correcting knowledge graphs. Since MP2D constructs dialog flows using knowledge graphs, one can automatically generate dialogs using the newest version of the knowledge graph to create up-to-date dialogs.

Moreover, employing a knowledge graph for the automatic generation of dialogues confers the distinct advantage of readily producing conversations that are current and up-to-date. 
As information evolves over time, manually updating dialogues can be challenging due to their inherently unstructured nature. 
However, knowledge graphs are dynamic entities that expand and adapt over time, with considerable research dedicated to enhancing them~\cite{paulheim2017knowledge, cohen2023evaluating}. 
Since MP2D constructs dialogue flows using knowledge graphs, it becomes feasible to automatically generate dialogues using the latest version of the knowledge graph, ensuring that the dialogues remain current and relevant to time-variant information.

\subsection{Generate Questions}
\label{3.2}
%With the retrieved multi-passage $P = \{p_1^\dagger, R_1, p_2^\dagger,R_2, ... p_n^\dagger\}$ where $p_i^\dagger = \{s_1, s_2 ... s_{k_i}\}$,  a Passage-to-Dialog (P2D) model is utilized to auto-regressively generate questions for each sentence within the passage $p_i^\dagger$ as an answer, thus creating a per-passage dialog as shown in bottom of Figure 2. That is, for a given input passage $p_i^\dagger$, the output dialog from the P2D model is $P2D(p_i^\dagger) = \{(q_1,s_1), (q_2,s_2), ..., (q_{k_i}, s_{k_i})\}$, and thereafter, it naturally shifts to the next topic via the existing relation sentence $R_i$, repeating this process. Ultimately, the multi-passage $P = \{p_1^\dagger, R_1, p_2^\dagger, R_2 ... p_n^\dagger\}$ is transformed into a dialog that encapsulates natural topic shifts in the form $\{P2D(p_1^\dagger), R_1, P2D(p_2^\dagger), …\}$. Following this, a post-processing stage is completed to generate a dialog that includes natural topic shifts.

%Using the retrieved multi-passage $P=\{p_1^\dagger, R_1, p_2^\dagger,R_2, ... ,p_n^\dagger\}$, a Passage-to-Dialog (P2D) model is employed to auto-regressively generate questions for each sentence within the passage $p_i^\dagger = \{s_1, s_2 ... s_{k_i}\}$ as an answer, thereby creating a per-passage dialog, as illustrated at the bottom of Figure 2. 
Using the retrieved multi-passage \textit{MP}, a Passage-to-Dialogue (P2D) model is employed to auto-regressively generate questions for each sentence within the passage $p_i^\dagger = \{s_1, s_2 ... s_{k_i}\}$ as an answer, thereby creating a per-passage dialogue, as illustrated at the bottom of Figure~\ref{figure2}. 
In essence, for a given input passage $p_i^\dagger$, the output dialog from the P2D model is $D_i = \{(q_1,s_1), (q_2,s_2), …, (q_{k_i}, s_{k_i})\}$, where $q_j$ represents the generated question based on the answer $s_j$.
%Subsequently, it naturally transitions to the next topic via the existing relation sentence $R_i$, repeating this process. 
%Subsequently, it naturally transitions to the subsequent topic through $R_i$ and repeats this process iteratively.
Subsequently, it naturally transitions to the subsequent topic through $R_i$ while seamlessly incorporating questions $Q_{_{R_i}}$ about $R_i$ and repeats this process iteratively.
Ultimately, the multi-passage \textit{MP} $= \{p_1^\dagger, R_1, p_2^\dagger, ... ,p_n^\dagger\}$ is transformed into a dialogue that encapsulates natural topic shifts in the form $\{D_1, Q_{R_1}, R_1, D_2, ... , D_n\}$.
%Following this, a post-processing stage that gives meta-information regarding the entire passage and topic shift turns is carried out to generate a dialogue that incorporates natural topic shifts.

%We adopt a Large Language Model (LLM) as the question generator for MP2D. This is based on observations from comparing datasets generated by various P2D models and LLMs as question generators, indicating that LLMs have the potential to better generate questions for topic-shift turns (Section 4.1). To employ LLMs as P2D generators, one naive approach could be to perform a dialog reconstruction task by filling in [BLANK] without prepending a specific prompt. However, such a naive prompt does not guarantee the generation of contextually relevant questions for topic shift turns or thereafter. We discover that despite the answer containing information about a different topic due to a topic-shift, it generates questions about the previous topic without recognizing the shift (Section 6.1). Therefore, during the question generation process, additional instructions are provided to recognize each topic shift as it occurs. Detailed information about instructions can be found in the Appendix.

We employ an LLM as the question generator for MP2D.
%This choice is informed by observations derived from comparing datasets generated by various P2D models and LLMs as question generators, which suggest that LLMs have the potential to more effectively generate questions for topic shift turns~(\S\ref{4.1}).
This decision is based on observations from comparing datasets generated by various P2D models and LLMs, which indicate that LLMs may generate questions for topic shift turns more effectively.
One possible approach to utilize LLMs as question generators is performing a dialogue reconstruction task by filling in [BLANK] without providing a specific prompt.
However, this approach does not ensure the generation of contextually relevant questions for topic shift turns or subsequent topics.
%We find that despite the answer containing information about a different topic due to a topic-shift, questions generated often pertain to the previous topic without recognizing the shift~(\S\ref{7}).
We often find that, despite the answer for the topic-shift turn including information about a new topic, the generated questions still pertain to the previous topic without recognizing the shift~(\S\ref{7}).
%Therefore, additional instruction is incorporated during the question-generation process to recognize each topic shift as it occurs.
Therefore, during the question-generation process, an additional instruction indicating a change in topic is provided in topic shift turns; \textit{"Note that the conversation topic has shifted to [next\_topic] from [current\_topic].}
%Prompts instruction and  can be found in the Appendix~\ref{detailed_MP2D}.
Detailed information, including the prompt for question generation and the post-processing steps, can be found in Appendix~\ref{detailed_MP2D}.
\section{Evaluating the MP2D Framework}
\label{4}
\begin{table*}[t]
\renewcommand{\arraystretch}{1.3}
\centering
\resizebox{0.9\textwidth}{!}{% 
\begin{tabular}{lcccccc}
\specialrule{1pt}{0pt}{0pt}  \hline
                 & \multicolumn{1}{l}{\textbf{USR-DR ($c$)}} & \multicolumn{1}{l}{\textbf{USR-DR ($f$)}} & \multicolumn{1}{l}{\textbf{  GPT2    }} & \multicolumn{1}{l}{\textbf{QRelScore\textsubscript{LRM}}} & \multicolumn{1}{l}{\textbf{QRelScore\textsubscript{GRG}}} & \multicolumn{1}{l}{\textbf{RQUGE}} \\ \hline
\multicolumn{3}{l}{\textbf{\textit{Single Passage}}}                                                     & \multicolumn{1}{l}{}                    & \multicolumn{1}{l}{}                      & \multicolumn{1}{l}{}                      & \multicolumn{1}{l}{}              \\ \hline 
Dialog Inpainter~\cite{dai2022dialog} & 0.9615                              & 0.7227                                   & 0.5125                                   & 0.4887                                     & 0.4808                                     & 3.1255                             \\
Dialogizer~\cite{hwang2023dialogizer}       & 0.9641                              & 0.7883                                   & 0.5386                                   & 0.5044                                     & 0.4852                                     & 3.2511                             \\
GPT-3.5    & \textbf{0.9856}                              & \textbf{0.8960}                                   &      \textbf{0.5739}                              & \textbf{0.5369}                                    & \textbf{0.5305}                                     & \textbf{3.2923}                             \\  \specialrule{1pt}{0pt}{0pt} 
\multicolumn{3}{l}{\textbf{\textit{Multiple Passages}}}                                                   & \multicolumn{1}{l}{}                    & \multicolumn{1}{l}{}                      & \multicolumn{1}{l}{}                      & \multicolumn{1}{l}{}              \\ \hline 
Dialog Inpainter~\cite{dai2022dialog} & 0.9389                              & 0.7160                                   & 0.4972                                   & 0.4874                                     & 0.4732                                     & 2.9156                             \\
   Dialogizer~\cite{hwang2023dialogizer}       & 0.9474                              & 0.7738                                   & 0.5034                                   & 0.5098                                     & 0.4748                                     & 2.9363                             \\
  GPT-3.5 (MP2D)   &  \textbf{0.9873}                              & \textbf{0.9199}                                   & \textbf{0.5746}                                   & \textbf{0.5366}                                     & \textbf{0.5437}                                     & \textbf{3.1034}                             \\ \hline \specialrule{1pt}{0pt}{0pt} 
\end{tabular}
 }
\caption{Automatic evaluation results obtained by assessing the generated dialogues using reference-free metrics.}
%MP2D (Ours) refers to our proposed framework, which adapts GPT-3.5 as a question generator.} 
\label{table1}
\vspace{-3mm}
\end{table*}

%In this section, we empirically demonstrate the high-quality capabilities of the MP2D framework in automatically generating topic-shifted dialogs, both quantitatively (section 4.1) and qualitatively(section 4.2). First, we employ existing passage-to-dialog methods and the LLM as question generators to compare these approaches using various reference-free dialog metrics. For evaluation purposes, we utilize the Wikipedia dataset to create 10k multi-turn dialogs using the MP2D framework's passage retrieval component and then assess their quality. Furthermore, absolute evaluations are conducted using various criteria through human and GPT-4 assessments to demonstrate that the generated topic shift dialogs exhibit a natural transition of topics.

%In this section, we empirically demonstrate the high-quality capabilities of the MP2D framework in automatically generating dialog with smooth topic shifts, both quantitatively(\S\ref{4.1}) and qualitatively(\S\ref{4.2}).
In this section, we empirically demonstrate that the MP2D framework is capable of automatically generating high-quality dialogue with smooth topic shifts, both quantitatively~(\S\ref{4.1}) and qualitatively~(\S\ref{4.2}).
First, we employ existing P2D methods and the LLM as question generators to compare these approaches using various reference-free dialogue metrics.
%We utilize the Wikipedia dataset to create 10,000 multi-turn dialogues using the MP2D framework's passage retrieval component, and subsequently, we assess their quality.
Within the MP2D framework's passage retrieval component, we employ the Wikipedia dataset to generate 10,000 multi-turn dialogues, subsequently evaluating their quality.
Furthermore, absolute evaluations are conducted using various criteria to illustrate that the generated topic shift dialogues exhibit a natural transition of topics.

\subsection{Automatic Evaluation}
\label{4.1}

\paragraph{Passage to Dialogue Methods}
%We conduct experiments comparing Dialog Inpainter, Dialogizer, and GPT-3.5 as question generation methods that generate questions autoregressively using retrieved passages as answers. For a fair comparison, we implement Dialog Inpainter and Dialogizer to align with the specific requirements of each framework. Both models employ T5-base as their backbone and are trained using four dialog datasets for training: Daily Dialog, Task Masker, OR-QUAC, and QReCC. For GPT-3.5, we utilize the gpt-3.5-turbo model and incorporate the instruction for topic shift turns, as described in Section 3.2, to generate questions.

%First, we perform a basic comparison of the question generation performance for single passages by applying these three question generators to randomly retrieved passages from Wikipedia. Additionally, in the context of multi-passage content generated by the MP2D framework, we compare the three models by using them to transform the multi-passage content into dialogs within the multi-passage setting.

%We conduct experiments comparing Dialog Inpainter, Dialogizer, and LLM as question generation methods that auto-regressively generate questions using retrieved passages as answers.
We conduct experiments to compare Dialog Inpainter~\cite{dai2022dialog}, Dialogizer~\cite{hwang2023dialogizer}, and the LLM as question generators.
To ensure a fair comparison, we implement Dialog Inpainter and Dialogizer to align with the specific framework requirements.
Both models utilize T5-base~\cite{raffel2020exploring} as their backbone and are trained with four datasets: Task Masker~\cite{byrne2019taskmaster}, Daily Dialog~\cite{li2017dailydialog}, OR-QUAC~\cite{qu2020open}, and QReCC~\cite{anantha2020open}.
For LLM, we employ GPT-3.5 and integrate the instructions for topic shift turns to generate questions.

First, we conduct a basic comparison of the question generation performance for single passages by applying these three question generators to randomly retrieved passages from Wikipedia.
Additionally, we compare the three models by employing them to transform the multi-passage content into dialogues.

\paragraph{Evaluation Metrics}

%Due to the one-to-many nature of dialog systems, reference-free metrics are known to exhibit a higher correlation with human judgment compared to reference-based generation metrics. Additionally, as MP2D serves as a data generation framework, we utilize various reference-free metrics to assess multiple aspects of dialog and ConvQA, aiming to quantitatively measure the performance of MP2D. Firstly, USR-DR is a reference-free dialog metric developed to assess dialog in terms of context maintenance, interest, and knowledge utilization. Focusing on context maintenance, we employ USR-DR(c), a metric that evaluates dialogs based on history and facts as inputs, and USR-DR(f), which assesses dialogs using fact information or context as inputs. The authors have introduced a metric based on the GPT-2 model that evaluates dialogs on context, fluency, diversity, and logical consistency. We utilize the context metric for assessing the coherence between sentences in a dialog. RQUGE is a question generation metric developed to evaluate the answerability of the question given the context. Similarly, Leo introduced QRelScore, a metric designed for evaluating context-aware question generation without the need for additional training or human supervision. QRelScore is divided into QRelScoreLRM and QRelScoreGRG; QRelScoreLRM evaluates complex reasoning through word-level similarity analysis, whereas QRelScoreGRG assesses factual accuracy by examining the confidence in generating contextually relevant content.

Given the one-to-many nature of dialogue systems~\cite{zhao2017learning}, reference-free metrics are acknowledged for their stronger correlation with human judgment compared to reference-based generation metrics~\cite{gupta2019investigating, zhang2021dynaeval} when assessing dialogue quality.
Furthermore, as MP2D functions as a data generation framework, we utilize various reference-free metrics to evaluate multiple aspects of dialogue and ConvQA to assess the performance of MP2D quantitatively.
Primarily, \textbf{USR-DR}~\cite{mehri2020usr} is a reference-free dialogue metric for evaluating dialogues on context maintenance, interest, and knowledge utilization.
We utilize USR-DR(c), a metric that evaluates dialogues based on history and facts as inputs, along with USR-DR(f), which assesses dialogues using fact information or context as inputs.
%For specific focus on context, we utilize USR-DR(c), a metric that evaluates dialogs based on history and facts as inputs, along with USR-DR(f), which assesses dialogs using fact information or context as inputs. %Additionally, \textbf{GPT-2} based  metric~\cite{pang2020towards} evaluates dialogs on context, fluency, diversity, and logical consistency, with the context metric used to assess the coherence between sentences. 
Additionally, \textbf{GPT-2} based metric~\cite{pang2020towards} evaluates dialogs to assess the coherence between utterances. 
Furthermore, \textbf{RQUGE}~\cite{mohammadshahi2022rquge} assesses question answerability given the context, while \textbf{QRelScore}~\cite{wang2022qrelscore} evaluates context-aware question generation without extra training or human supervision. QRelScore is divided into QRelScore\textsubscript{LRM}, which evaluates complex reasoning through word-level similarity analysis, and QRelScore\textsubscript{GRG}, assessing factual accuracy by examining the confidence in generating contextually relevant content.
%QRelScore\textsubscript{LRM} evaluates complex reasoning through word-level similarity analysis, whereas QRelScore\textsubscript{GRG} assesses factual accuracy by examining the confidence in generating contextually relevant content.

\paragraph{Results}

%Table 1 presents the comparative results of three question generator methods within single passage and multi-passage settings. First, the single passage experiment aims to compare the performance of question generators in a basic single passage context by randomly selecting an entity without utilizing the paths of a knowledge graph and retrieving a passage accordingly. The experimental results reveal that using the GPT-3.5 for question generation outperforms both dialog inpainter and dialogizer across all metrics.

Table~\ref{table1} compares three question generation methods in single- and multi-passage settings.
In the single passage setting, we aim to evaluate the performance of question generators within a basic single passage context by randomly selecting an entity without the assistance of knowledge graph pathways for passage retrieval.
The results indicate that using GPT-3.5 for question generation outperforms both Dialog Inpainter and Dialogizer across all metrics.

Moving to the multi-passage setting, which introduces additional complexity due to the need to manage topic transitions and generate responses that effectively bridge these shifts, GPT-3.5 consistently exhibits superior performance.
Metrics assessing the dialogue's overall context or the relevance of question-answer (QA) pairs indicate a performance decline in finetuned models relative to the single-passage setting.
%When considering metrics that assess the overall context of the dialog or the overall question-answer relevancy, finetuned models show a decrease in performance compared to the single-passage setting. 
%Conversely, GPT-3.5-generated datasets maintain high performance levels even in multi-passage settings, suggesting its robustness in handling more intricate dialog generation tasks involving multiple passages.
Conversely, GPT-3.5-generated datasets maintain high performance even in multi-passage settings, suggesting its robustness in handling more complex question-generation tasks.
%However, in case of the RQUGE, which measure the contextual relevance between the given passage and target QA pair, decline in all methods when applied to multi-passage contexts.
However, in the case of the RQUGE, which measures the contextual relevance between the given passage and target QA pair, there is a decline in all methods when applied to multi-passage contexts.
This reduction is attributed to the broader range of topics covered by multi-passage content as opposed to single-passage content, which leads to a lower relevancy score between the multi-passage content and an individual QA pair. 

In summary, when creating dialogues from multi-passage content, LLM proves to be more effective as a question generator than task-specific finetuned models. 
%Despite observing a minor reduction in score for single QA pair evaluation within multi-passage context, MP2D demonstrates a robust ability to generate dialogues of comparable quality to those created from a single passage, while also incorporating topic shifts within the dialog framework.
%Despite observing a minor reduction in score for single QA pair evaluations within a multi-passage context, MP2D demonstrates a robust ability to generate dialogs of comparable quality to those created from a single passage, while also incorporating topic shifts within the dialog.
%Despite observing a minor reduction in the score for single QA pair evaluations within a multi-passage context, MP2D demonstrates a robust ability to generate dialogues of comparable quality to those created from a single passage while also incorporating topic shifts within the dialogue.
Furthermore, MP2D demonstrates a robust ability to generate dialogues of comparable quality to those created from a single passage while incorporating topic shifts within the dialogue.

%In summary, when creating dialogs from multi-passage content, LLM proves to be more effective as a question generator than task-specific finetuned models. Although there is a slight drop in scores when evaluating questions and answers within the context of passages, MP2D demonstrates its ability to generate dialogs of comparable quality to those created from single passage to dialog, while also incorporating topic shifts within the dialog framework.

\begin{table}[t]
\renewcommand{\arraystretch}{1.3} 
\centering
\resizebox{0.95\columnwidth}{!}{
\begin{tabular}{llllcc}
\hline \hline
\multicolumn{4}{c}{}                               & \textbf{Human}        & \textbf{GPT-4}        \\ \hline
\multicolumn{2}{l}{\textbf{Topic shift timing}}  &         &         & \multirow{2}{*}{95.67\%} & \multirow{2}{*}{93\%} \\
\multicolumn{4}{l}{\textit{Is the timing of topic-shifts is natural?}}        &                       &                       \\ \hline
\multicolumn{2}{l}{\textbf{Topic shift fluency}} &         &         & \multirow{2}{*}{87.67\%} & \multirow{2}{*}{88\%} \\
\multicolumn{4}{l}{\textit{Does the topic shifts occur smoothly?}}            &                       &                       \\ \hline
\multicolumn{2}{l}{\textbf{Overall quality}}     &         &         & \multirow{2}{*}{84.33\%} & \multirow{2}{*}{89\%} \\
\multicolumn{4}{l}{\textit{Is the overall quality of the dialog is good?}}    &                       &                       \\ \hline
\multicolumn{2}{l}{\textbf{Toxicity}}            &         &         & \multirow{2}{*}{\underbar{0\%}} & \multirow{2}{*}{\underbar{0\%}} \\
\multicolumn{4}{l}{\textit{Does the whole text has any potential risk?}}      &                       &                       \\ \hline \hline
\end{tabular}
}
\caption{The human and GPT-4 evaluation results.}
\label{table2}
\vspace{-3mm}
\end{table}

\subsection{Human \& GPT-4 Evaluation}
\label{4.2}
%To qualitatively assess the quality of topic shift dialog data generated by MP2D, we employ both human and GPT-4 as evaluators. GPT-3.5 is utilized as the question generator for MP2D, and from the 10k multi-turn dialogs previously generated, 100 sample dialogs are randomly selected to create the evaluation dataset. For evaluation, three criteria are used: the timing of the topic shift, the naturalness of the topic shift, and the overall quality of the dialog. The timing of the topic shift evaluates whether the topic changes at an appropriate moment as the conversation progresses within the dialog. The naturalness of the topic shift assesses whether the utterances involved in the topic shift are seamlessly and naturally connected. Lastly, the overall quality measures the excellence of the entire dialog, including the incorporation of the topic shift.

To qualitatively assess the quality of topic shift dialogue data generated by MP2D, we employ human and GPT-4~\cite{openai2023gpt} as evaluators. 
GPT-3.5 serves as the question generator for MP2D, and from the previously generated 10,000 dialogues, we randomly select 100 sample dialogues to form the evaluation dataset.
The evaluation focuses on three criteria: the timing of the topic shift, the naturalness of the topic shift, and the overall quality of the dialogue. 

%The timing criterion examines whether the topic changes at an appropriate moment as the conversation progresses within the dialog. The naturalness criterion assesses the seamless and logical connection of utterances during a topic shift. The overall quality criterion evaluates the dialog's excellence, including how well topic shifts are integrated.

%As indicated in Table 2, human evaluation results demonstrate that over 90\% of the MP2D-generated topic shift dataset exhibits topic shifts at appropriate timings, more than 80\% of the dialogs have smooth topic transitions, and the overall quality of over 90\% of the dialogs is assessed as excellent. The experimental results using GPT-4 as an evaluator also show a similar trend to that of the human evaluation. Upon investigating the reasons why some dialogs are rated as having unnatural topic shifts, we find that instances of unnaturalness are often attributed to paths in the knowledge graph where there is a rapid transition between entities from specific to general levels. A detailed case study on this matter can be observed in Section 5. In addition to the three evaluation criteria, we further ensure the dataset is devoid of toxicity by investigating all texts within the dataset for potential toxicity; it is confirmed that no dialog contains toxicity.

Table~\ref{table2} presents the human evaluation results, indicating that over 95.6\% of the MP2D-generated topic shift dataset exhibits timely topic shifts, with more than 87.6\% of the dialogues demonstrating smooth topic transitions, and over 84.3\% rated as excellent in overall quality. 
Moreover, experimental results using GPT-4 as an evaluator~\cite{liu2023gpteval,gilardi2023chatgpt} reflect a similar trend to that of the human evaluation.
An analysis of dialogues rated with unnatural topic shifts reveals that such instances often arise from abrupt transitions between entities at different levels of specificity within the knowledge graph. 
A comprehensive case study on this observation is provided in Section~\ref{7}. 
Beyond the primary evaluation criteria, we also screen the evaluation dataset for toxicity and confirm the absence of toxic content in any dialogues. 
More detailed information, including inter-annotator agreement, compensation details, instructions, and prompts, can be found in Appendices ~\ref{human evaluation} and ~\ref{gpt-4 evaluation}.

\section{Topic Shift Dialogue Benchmark}
\label{5}

\subsection{TS-WikiDialog}

\label{5.1}
%We introduce the novel topic shift ConvQA benchmark TS-WikiDiag to evaluate the performance of LLMs across various topic shift tasks and to demonstrate the utility of datasets generated by the MP2D framework through application experiments. The data generated through the passage to dialog method, where each answer sequentially spans the passage, cannot be directly applied to tasks such as the ConvQA response generation due to its structured format. Therefore, we paraphrase the answer portions of the dialogs generated by the MP2D framework to adapt the benchmark to the ConvQA response generation task. The paraphrase process involves providing GPT-4 with the dialog history and target answer for paraphrasing, followed by manual adjustments by humans to ensure the context is naturally preserved. Furthermore, to ensure a fair evaluation, TS-WikiDiag was designed to exclude any overlapping entities with the dialogs generated for experiments in section 4 and 6. TS-WikiDiag encompasses 1k multi-turn dialogs, with a total of 39k turns. Each dialog approximately covers 2.2 topics, and more detailed statistical analysis can be observed in Table 3.

%We introduce TS-WikiDialog, a benchmark for evaluating LLMs on topic shift tasks and illustrating the usefulness of MP2D. 
We introduce TS-WikiDialog, a benchmark designed to evaluate LLMs on topic shift tasks and to illustrate the usefulness of MP2D.
%The data generated via the P2D methods, where each answer sequentially spans the passage, cannot be directly applied to tasks such as ConvQA response generation due to its structured format. 
The data generated by the P2D methods cannot be directly utilized for tasks such as ConvQA response generation because each answer sequentially spans across the passage.
%Therefore, we paraphrase the answers of the dialogues generated by the MP2D framework to adapt the benchmark to the ConvQA response generation task. 
Therefore, to adapt the benchmark to the ConvQA response generation task, we paraphrase the answers in the MP2D-generated dialogues accordingly.
The paraphrasing process involves providing GPT-4 with the dialogue history and target answer for paraphrasing, followed by manual adjustments by humans to ensure the context is naturally preserved. 
%Furthermore, the few instances of unnatural topic flow are filtered or revised to ensure a more natural progression.
Furthermore, the few instances of unnatural topic flow present in the automatically generated data are filtered or revised to ensure a more natural progression.
%In addition, to ensure a fair evaluation, TS-WikiDiag is designed to exclude any overlapping entities with the dialogs generated for experiments in Section~\S\ref{4}. 
TS-WikiDialog encompasses 1,000 multi-turn dialogues, totaling 15,892 turns. 
Each dialogue covers approximately 2.136 topics, and more detailed statistical analysis can be observed in Appendix~\ref{datasetstatistic}.

\begin{figure}[t]
\centering
\includegraphics[width= 1\columnwidth]{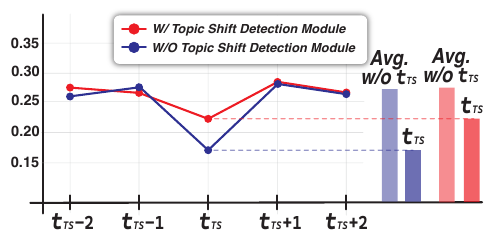} 
\caption{Results of the ConvQA response generation performance of GPT-3.5. Each score represents the BLEU-4 score, where t\textsubscript{TS} denotes a topic shift turn.}
\label{figure3}
\vspace{-3mm}
\end{figure}

\subsection{The Struggle of LLM in Topic Shift turns}
Using TS-WikiDialog, we assess how well the LLM maintains its response generation quality in the face of topic transitions. 
%In setting up the experiment for ConvQA response generation, we provide multi-passage content, dialog history, and the target question to GPT-3.5, and measured the answers generated by the model.
For the ConvQA response generation experiment, we set multi-passage content, dialogue history, and the target question as the input for the GPT-3.5 and evaluate the answers generated by the model. 
%The findings, as depicted in Figure~\ref{figure3}, indicate deterioration in the performance of GPT-3.5 during instances of topic shift, emphasizing the imperative for even LLMs to enhance their proficiency in navigating topic transitions. 
%The findings, as depicted in Figure~\ref{figure3} (blue line), indicate a deterioration in the performance of GPT-3.5 on topic shift turns, emphasizing that even LLMs encounter challenges in handling topic transitions.
The results can be observed in Figure~\ref{figure3} (blue line), where each score represents the BLEU-4 score~\cite{papineni2002bleu} between the candidate answer and the reference. 
We observe a decrease in the performance of GPT-3.5 in topic shift turns ($t_{_{TS}}$), emphasizing that even LLMs face challenges in managing topic transitions.

%~\cite{papineni2002bleu} between the candidate answer and the reference

%In Section~\S\ref{6.3}, we demonstrate the potential for improvement by employing a model fine-tuned with an MP2D-generated dataset, without additional LLM training.

%definition sentence는 성능이 높고, topic shift turn은 성능이 낮음

\begin{table}[t]
\renewcommand{\arraystretch}{1.25} 
\centering
\resizebox{0.9\columnwidth}{!}{
\begin{tabular}{lcccc}
\hline \hline
\multicolumn{1}{c}{\textit{}} &\textbf{ \textit{F1}}     & \textbf{\textit{P}}     & \textbf{\textit{R}}    & \textbf{\textit{EM}}    \\ \hline
\multicolumn{5}{l}{\textit{Zero-shot (w/o In-Context Learning)}}                                                  \\ \hline 
GPT-3.5 (175B)                & 0.303           & 0.548           & 0.227          & 0.000          \\
GPT-4 (>175B)                  & 0.498           & 0.827           & 0.448           & 0.000          \\ \hline
\multicolumn{5}{l}{\textit{Few-shot (w/ In-Context Learning)}}                                                    \\ \hline
GPT-3.5 (175B)                & 0.524           & 0.826           & 0.487          & 0.000          \\
GPT-4 (>175B)                  & 0.524           & 0.841           & 0.503           & 0.000          \\ \hline
\multicolumn{5}{l}{\textit{MP2D Random-shift}}                                                    \\ \hline
T5-base (220M)                & 0.867           & 0.921           & 0.873          & 0.385          \\
FLAN-T5-base (220M)           & 0.879           & 0.928           & 0.887           & 0.441          \\ \hline
\multicolumn{5}{l}{\textbf{\textit{MP2D Knowledge-Graph (Ours)}}}                                           \\ \hline
T5-base (220M)                & \textbf{0.970 } & \underbar{0.987}           & \underbar{0.973}          & \underbar{0.908}          \\
FLAN-T5-base (220M)           & \textbf{0.970 } & \textbf{0.988 } & \textbf{0.975} & \textbf{0.913} \\ \hline \hline
\end{tabular}
}
\caption{The results of Topic Segmentation Task}
\label{table4}
\vspace{-3.5mm}
\end{table}

\section{Applications}
%=To illustrate the efficacy of MP2D, we evaluate the performance of models trained with datasets generated by the framework against the zero-shot capabilities of LLMs across various topic shift dialogue tasks. 
%To illustrate the efficacy of MP2D, we evaluate the performance of models trained on datasets produced by MP2D framework against the zero-shot and few-shot capabilities of LLMs in two topic shift tasks: topic segmentation and topic shift detection.
To illustrate the efficacy of MP2D, we evaluate the performance of models trained on datasets produced by the MP2D framework in two topic shift tasks: topic segmentation and topic shift detection.
Since MP2D is the first framework for automatically generating topic shift ConvQA data, we experiment with baseline models that include zero-shot and few-shot settings of LLMs, as well as models trained on dialogues constructed from randomly selected topics without the use of knowledge graphs (\textit{MP2D Random-shift}).
%Our analysis aims to validate that the MP2D framework proficiently generates datasets that train models to achieve high performance on diverse topic shift tasks, establishing its value as a dataset generation tool. 
Our analysis aims to validate that the MP2D framework is an effective data generation tool, producing datasets that enable models to achieve high performance on diverse topic shift tasks.
%To prevent overlaps in topics or retrieved passages between training and test datasets, we employ the 10,000 dialogue dataset generated in Section~\ref{4} for training and utilize the TS-WikiDiag 1k dataset for testing. 
We generate 10,000 dialogues using the MP2D framework for training purposes and employ TS-WikiDialog as the test set, ensuring there is no overlap in topics or retrieved passages between the training and test sets.
%We assess T5-base and FLAN-T5-base, trained on MP2D-generated datasets, against GPT-3.5 and GPT-4 in zero-shot and few-shot settings, reflecting the absence or presence of In-context Learning (ICL).
Further details, including the prompts for each task, can be found in Appendix~\ref{topic_shift_dialog_tasks_details}.

%실험 세팅 알아듣기 쉽게
%diag 별로 다시 합친 내용 (어떻게 실험했는지 쉽게 읽히게끔)
%Metric 들도 알아보기 쉽게

%[반드시 언급해야할 내용]
%EM 0.000
%detection 에서 GPT-4 가 못하는 이유 설명

\subsection{Topic Segmentation}
\label{6.1}
The objective of topic segmentation is to partition the given dialogue into segments based on the topics being discussed~\cite{arguello2006topic}. 
Given a dialogue $D$ composed of utterances $\{u_1, u_2, ..., u_n\}$, where $u_i$ represents an utterance, the model's output would be a sequence of segment labels, for example, $\{0, 0, 1, 1, ..., 2\}$, with each numeral indicating a distinct topic segment. 
The topic segmentation task has the potential to enhance the performance of tasks such as information retrieval and dialogue summarization by facilitating the identification of topic boundaries within a conversation~\cite{lin2023topic, gao2023unsupervised}.

%We assess the performance of finetuned Flan-T5 models trained with MP2D against the zero-shot and few-shot capabilities of various LLMs. 
We measure the F1, Precision (P), Recall (R), Exact Match (EM).
EM is required to delineate topics across all utterances within a dialogue precisely.
%The experimental results can be shown in Table~\ref{table4}. 
%GPT-3.5 (175B) and GPT-4 (>175B), exhibit poor performance in both settings in all metrics.
GPT-3.5 (175B) and GPT-4 (>175B) exhibit poor performance in all metrics for both zero-shot and few-shot settings, as shown in Table~\ref{table4}.
% (>175B\footnote{This is an estimated parameter size.})
Notably, they fail to achieve an Exact Match (EM) in all dialogues, demonstrating the considerable challenge of identifying topics within dialogues. 
%In contrast, T5-base and Flan-T5-base, with only 220M parameters, prove that training on datasets generated by the MP2D method can learn topic shifts, evidenced by high F1 scores of 0.87 (\textit{Random-shift}) and 0.97 (\textit{Knowledge-graph}).
%Furthermore, training with MP2D using \textit{Knowledge-graph} shows better performance in both EM and F1 than training with \textit{Random-shift}, validating our proposed method that leverages knowledge graphs to generate dialogues reflecting natural topic shifts.
In contrast, T5 and Flan-T5 models (220M), when finetuned on MP2D-generated datasets, exhibit superior performance across all metrics. 
Furthermore, models trained on \textit{Knowledge-graph} outperform those trained on \textit{Random-shift}, proving the effectiveness of our proposed method that leverages knowledge graphs in generating natural topic shifts.

\begin{table}[t]
\renewcommand{\arraystretch}{1.4} 
\centering
\resizebox{1\columnwidth}{!}{
\begin{tabular}{lccccc}
\hline \hline
\multicolumn{1}{c}{\textit{}} & \textit{\textbf{Acc.}}                   & \textit{\textbf{F1}} & \textit{\textbf{P}}  & \textit{\textbf{R}}  & \textit{\textbf{EM}} \\ \addlinespace[-8pt]
                              & \multicolumn{1}{l}{\textit{{\small(per turns)}}} & \multicolumn{1}{l}{} & \multicolumn{1}{l}{} & \multicolumn{1}{l}{} & \multicolumn{1}{l}{} \\ \hline
\textit{Zero-shot (w/o In-Context Learning)}  &                                          &                      & \multicolumn{1}{l}{} & \multicolumn{1}{l}{} &                      \\ \hline
GPT-3.5 (175B)                & 0.720                                    & 0.701                & 0.751                & 0.719                & 0.000                \\
GPT-4 (>175B)                  & 0.254                                    & 0.222                & 0.826                & 0.277                & 0.000                \\ \hline
\multicolumn{6}{l}{\textit{Few-shot (w/ In-Context Learning)}}                                                                                                                       \\ \hline
GPT-3.5 (175B)                & 0.757                                    & 0.713                & 0.784                & 0.741                & 0.000                \\
GPT-4 (>175B)                  & 0.275                                    & 0.246                & 0.780                & 0.283                & 0.000                \\ \hline
\multicolumn{6}{l}{\textit{MP2D Random-shift}}                                                                                                                       \\ \hline
T5-base (220M)                & 0.825                                    & 0.739                & 0.843                & 0.809                & 0.016                \\
FLAN-T5-base (220M)           & 0.824                                    & 0.741                & 0.841                & 0.808                & 0.035                \\ \hline
\multicolumn{6}{l}{\textit{\textbf{MP2D Knowledge-Graph (Ours)}}}                                                                                                    \\ \hline
T5-base (220M)                & \textbf{0.841}                           & \textbf{0.803}       & \textbf{0.885}       & \textbf{0.834}       & \textbf{0.225}       \\
FLAN-T5-base (220M)           & \textbf{0.841}                           & \underbar{0.792}                & \underbar{0.873}                & \underbar{0.833}                & \underbar{0.205}                \\ \hline \hline
\end{tabular}
 }
\caption{The results of Topic Shift Detection Task}
\label{table5}
\vspace{-3.5mm}
\end{table}

\begin{table*}[ht]
\renewcommand{\arraystretch}{1.03}
\centering
\resizebox{0.82\textwidth}{!}{% 
\begin{tabular}{l|ll}
\hline \hline
                        & \hspace{3.5mm}Q: &       What was \textit{Lekain}'s education and how did it contribute to his early career as an actor?            \\
                        & \hspace{3.5mm}A: &      He was educated at the Collège Mazarin, and joined an amateur company of players against which         \\
                        &    &        the Comédie-Française obtained an injunction.                                               \\
\hspace{-0.05em}\raisebox{-0.5ex}{\begin{important_blue}\textbf{Case 1}\end{important_blue}}                  & \hspace{3.5mm}Q: &   Was there no student of \textit{Lekain}?                                                                   \\
                        & \hspace{3.5mm}A: &      \textit{Larive} was a student of \textit{Lekain}.                                                                   \\
                        & \hspace{3.5mm}Q: &        What can you tell me about \textit{Larive}?                                                                \\
                        & \hspace{3.5mm}A: &        Jean Mauduit, stage name \textit{Larive} or de La Rive was a French actor.                                 \\ \hline
                        & \hspace{3.5mm}Q: &       What is the geological age of \textit{Rhacheosaurus}?                                                      \\
&\hspace{3.5mm}A: &        The genus \textit{Rhacheosaurus} is a fossil taxon, belonging to the \textit{Metriorhynchidae} family.              \\ \multirow{2}{*}{\hspace{-0.05em}\raisebox{-0.5ex}{\begin{important_red} \textbf{Case 2}\end{important_red}}} 
                        & \textcolor{red}{\textcolor{red!100!black}{\xmark} \hspace{0.5mm}Q:} &         \textcolor{red}{What is the geographic distribution of \textit{Rhacheosaurus?}}                                            \\
                        &\textcolor{blue!80!black}{\cmark} Q: &        What is \textit{Metriorhynchidae}, and during which geological periods and in which regions did it exist? \\
                        & \hspace{3.5mm}A: &          \textit{Metriorhynchidae} is an extinct family of specialized, aquatic metriorhynchoid crocodyliforms from \\
                        &    &         the Middle Jurassic to the Early Cretaceous period of Europe, North America and South America.    \\ \hline
                        & \hspace{3.5mm}Q: &        When was the \textit{Malcolm Group} founded?                                                               \\
                        & \hspace{3.5mm}A: &        \textit{Malcolm Group}, a \textit{logistics} business founded in 1960, is located in Linwood, Renfrewshire.         \\
\hspace{-0.05em}\raisebox{-0.5ex}{\begin{important_red}\textbf{Case 3}\end{important_red}}                  &\hspace{3.5mm}Q:       & What is the definition of \textit{logistics}?                                                              \\
                        & \hspace{3.5mm}A: & \textit{Logistics} is the part of supply chain management that deals with the efficient forward and reverse           \\
                        &    &   flow of goods, services, and related information from the point of origin to ...      \\ \hline \hline
\end{tabular}
}
%\caption{Case Study. Question marked in \textcolor{red}{red} is generated without explicit instruction indicating a topic shift turn.} 
\caption{Case Study. \textbf{Case 1}: A successful example. \textbf{Case 2}: An example of inaccurate question generation from lacking additional instruction in a topic shift turn. The question marked in \textcolor{red}{red} is generated without the instruction. \textbf{Case 3}: An example that might seem unnatural due to an abrupt change from specific to general topics.}
\label{table6}
\vspace{-4mm}
\end{table*}
\subsection{Topic Shift Detection}
\label{6.2}
%Topic-Shift Detection is a task to identify instances within a conversation where the subject undergoes a real-time change~\cite{lin2023multi}. 
%Topic-Shift Detection is a task designed to identify instances within a conversation where the subject undergoes a change in real-time~\cite{lin2023multi}.
Topic Shift Detection is the task of detecting topic transitions of a conversation in real-time~\cite{lin2023multi}.
This task is crucial for real-time dialogue systems, empowering them to adapt dynamically to the evolving conversation, thereby facilitating relevant responses based on the detected topic shifts. 
%As the dialogue unfolds and each utterance is sequentially added, the objective is to determine whether a given utterance signifies a continuation of the ongoing topic or marks the inception of a fresh one. 
As the dialogue progresses with each sequential utterance, the objective is to identify whether a target utterance continues the current topic or initiates a new one.
%For example, given the sequence $\{u_1, u_2, ..., u_i\}$, if the last utterance $(u_i)$ introduces a shift in topic, the system would produce a '1' to signify this change. Conversely, if $u_i$ maintains the current topic, the output would be '0'. 
For instance, with the sequence $\{u_1, u_2, ..., u_i\}$, the system outputs `1' if the last utterance $u_i$ introduces a shift in topic; otherwise, it produces a `0' if $u_i$ continues the current topic.

%Similar to the topic segmentation experiment, we conduct a comparison involving our Flan-T5 model, trained with MP2D, against the zero-shot and few-shot performance of various LLMs. 
Table~\ref{table5} presents results consistent with those discussed in Section~\ref{6.1}, demonstrating that models finetuned on MP2D-generated datasets uniformly surpass the baselines in all metrics. 
%LLMs fail in all cases for EM and show a similar trend in accuracy per turn, which assesses whether a topic shift occurs at each utterance turn. 
LLMs fail in all cases for EM and show a similar trend in accuracy per turn, which evaluates the correctness of detecting a topic shift at each utterance turn.
Notably, GPT-4 exhibits significantly lower results compared to GPT-3.5 due to its sensitivity to topic changes and more detailed breakdown of topics.
%Notably, GPT-4 exhibits significantly lower results compared to GPT-3.5. This phenomenon can be attributed to the fact that GPT-4 delineates topics with a higher degree of granularity, thereby rendering it exceptionally responsive to the incidence of topic shifts.

\subsection{Enhancing LLM in Topic Shift turns}
\label{6.3}

%We enhance the ConvQA response generation performance of LLMs specifically at topic shift turns by integrating the fine-tuned Flan-T5 model with MP2D-generated data for topic shift detection. 
%We enhance the ConvQA response generation performance of LLMs specifically at topic shift turns by integrating the topic shift detection module.
We enhance the ConvQA response generation performance of LLMs, specifically at topic shift turns, by integrating the topic shift detection module. 
The finetuned Flan-T5 model is utilized as the detection module, which is trained on MP2D-generated data for topic shift detection tasks. 
Prior to feeding multi-passage content, dialogue history, and the target question into the LLM, they undergo processing through a topic shift detection model to ascertain if a topic shift has occurred in the current turn. 
This data is then integrated into the input for the LLM. 
Essentially, the LLM receives additional information regarding topic shifts to aid in generating the subsequent response. 
As depicted in Figure~\ref{figure3} (red line), the experimental findings validate that this information enhances the response generation performance at topic shift turns. 
Through this analysis, we illustrate the capability to enhance the performance of the LLM in handling topic shifts by utilizing data generated by MP2D without necessitating direct LLM training.

\section{Case Study}
\label{7}

%Examples drawn from the MP2D-generated topic shift dialogs are presented in Table~\ref{table6}. 
We present examples drawn from the MP2D-generated topic shift dialogues in Table~\ref{table6}.
Case 1 demonstrates a dialogue with a natural flow of the topic, and most dialogues generated by the MP2D framework include such topic transitions. %Case 2 serves as an example, as described in Section~\ref{3.2}, showcasing that when converting multi-passage content to ConvQA using GPT-3.5 without providing specific instructions at topic shift turns, it incorrectly generates questions about the previous topic. 
Case 2, described in Section~\ref{3.2}, demonstrates that converting multi-passage content to ConvQA with GPT-3.5 without clear instructions at topic shift turns leads to incorrect question generation. 
The question marked in red represents the question generated in a setting without instructions, illustrating that despite the topic shifting from \textit{Rhacheosaurus} to \textit{Metriorhynchidae} and the subsequent answer discussing the changed topic, it erroneously generates questions about \textit{Rhacheosaurus}. 
The question presented below the red-marked question is generated using the MP2D framework, successfully creating a question relevant to the changed topic.
%Case 3 is an example cited in Section~\ref{4.2} as exhibiting a challenging dialog flow. 
Case 3 is an example of a dialogue flow that presents challenges, as cited in Section~\ref{4.2}.
%The topic transitions from the \textit{Malcolm Group} to \textit{logistics}, although a relationship exists between the two, the abrupt change from a specific to a more general subject may be deemed unnatural.
The text transitions from discussing the \textit{Malcolm Group} to \textit{logistics}. Although there is a relationship between the two, the abrupt shift from a specific topic to a more general one might be considered unnatural.
Controlling the relationship between such entities during the path construction process presents a promising avenue for future work.
\section{Conclusion}
In this work, we address the data scarcity issue in topic shift dialogues by proposing a framework that automatically generates dialogues with natural topic transitions.
Our MP2D methodology utilizes the flow of relationships between entities in a knowledge graph to structure the dialogue flow and converts multi-passage content into a ConvQA format. 
Experimentally, we demonstrate that MP2D-generated topic shift dialogues are of high quality and prove their value as a training dataset for various topic shift dialogue tasks.

\section*{Limitations}
This study employs an LLM as the question generator within the multi-passage to dialogue framework, opting for its superior performance over the finetuned T5 models. However, when employed as a question generator, a performance-efficiency trade-off exists between T5 and LLMs; while T5 may exhibit slightly lower performance, it incurs less cost~\cite{patterson2021carbon}. Given that MP2D serves as a dataset generation framework, the inference cost can influence the size of the generated dataset. Therefore, when utilizing MP2D for dataset creation, it is crucial to make a context-appropriate judgment regarding the trade-off between quality and data size, allowing for selecting either a finetuned model or an LLM as the question generator.

The second limitation arises from the costs associated with LLMs~\cite{musser2023cost}, leading to the inability to employ GPT-4 in certain experiments due to financial constraints. For instance, when generating 10,000 dialogues in Section~\ref{4.1}, we used GPT-3.5 instead of GPT-4. It remains an open question whether employing GPT-4 to generate the dialogues could have resulted in higher performance for topic shift tasks.

Additionally, this study did not tackle the challenge of disambiguating entities in a knowledge graph, such as distinguishing between "Python," the programming language, and "Python," the snake on Wikipedia~\cite{chen2021evaluating}. In addressing the existence of multiple links for a single entity, our approach simply disambiguates by utilizing the first page returned in search results. Even though we observed only a small fraction of entities (approximately 1.34\%) exhibit such ambiguity, controlling the path to create more contextual topic shifts or to establish personalized topic flows represents a promising direction for future work.

\section*{Ethics Statement}

In Section~\ref{4.2}, we verify through crowd workers that the generated datasets are free of any potential ethical concerns. Such concerns include offensive, sexist, or racist remarks, toxic language, or any depictions of sexual behavior. The crowd workers received fair compensation for their evaluation of the dataset. A comprehensive description, the interface used for collecting human evaluations, and detailed information regarding compensation are provided in Appendix~\ref{human evaluation}.

Furthermore, we employ LLMs from the official website of OpenAI\footnote{\url{https://openai.com/}}. All models and datasets utilized in our research are sourced from publicly available websites or GitHub repositories.

% \newpage

\bibliography{custom}

\newpage
\clearpage

\appendix
\label{sec:appendix}
\section{Generated Datasets Statistics}
\label{datasetstatistic}
\begin{table}[h]
\renewcommand{\arraystretch}{1.3} 
\centering
\resizebox{0.95\columnwidth}{!}{
\begin{tabular}{ccc}
\hline \hline
                                   & \textbf{Section~\ref{4}} & \textbf{Section~\ref{5}} \\ \hline
\textbf{\textit{\# of dialogues}}             & 10,000            & 1,000             \\ \hline
\textbf{\textit{\# of turns}}               & 156,856                 & 15,892            \\ \hline
\textbf{\textit{Average \# of topics}}      & 2.137             & 2.136             \\ \hline
\textbf{\textit{Average tokens per dialogue}} & 22.50                & 23.26             \\ \hline
\textbf{\textit{\# of unique tokens}}       & 25,546                 & 15,938            \\ \hline \hline
\end{tabular}
}
\caption{The statistics of the generated datasets in Sections~\ref{4} and~\ref{5}.}
\label{table1_appendix}
%\vspace{-5mm}
\end{table}
The statistics for the multi-passage to dialogue datasets generated for automatic evaluation (Section~\ref{4}) and TS-WikiDialog (Section~\ref{5}) can be observed in Table~\ref{table1_appendix}. 

%In the creation of the dialog dataset, the single-passage experiments and the multi-passage experiments each utilized identical passages respectively.
\section{Reproducibility checklists}
\subsection{Dataset and Source code} 
Our experiment source code and configuration details are included as supplementary materials. The datasets produced and the complete codes, including weight parameters, will be made available to the public.

\subsection{Computing Resources}
For the experiments, Xeon 4210R (2.40 GHz) with RTX A6000 is employed. Four GPUs are utilized for the experimental setup.  All codes are implemented on Python 3.7.13 and PyTorch 1.10.1.

\subsection{Versions of the LLMs}
The GPT-3.5 version utilized for MP2D framework, topic segmentation, and topic shift detection is \textit{gpt-3.5-turbo-0613}. Moreover, the GPT-4 version employed for GPT-4 evaluation, ConvQA response generation, topic segmentation, and topic shift detection is \textit{gpt-4-0613}.

\section{Details of MP2D Implementation}
\label{detailed_MP2D}

In the question generation process of the MP2D framework, the prompts used when employing GPT-3.5 as the question generator are provided in Table~\ref{Table_dialog_generation_prompt}. 

The topic shift dialog created through the MP2D framework undergoes a simple post-processing step. Since questions are generated to fill in the [BLANK] in "A: [BLANK]," a few samples are produced with "A: " prefixed to the question, which was then removed through a rule-based approach.

\section{Details of P2D models Implementation}
In implementing the passage to dialogue methods(Dialog Inpainter and Dialogizer), we fundamentally adhere to the best-performing hyperparameters as proposed in their respective research for a fair comparison. The T5-base model\footnote{https://huggingface.co/t5-base} serves as the backbone for both models. To ensure the robustness of our findings, we conduct all experiments using three different seed numbers. Both models are trained with a batch size of 8 and a gradient accumulation step size of 8.

\section{Human Evaluation}
\label{human evaluation}

\begin{figure*}[t]
\centering
\fbox{\includegraphics[width=0.85\textwidth]{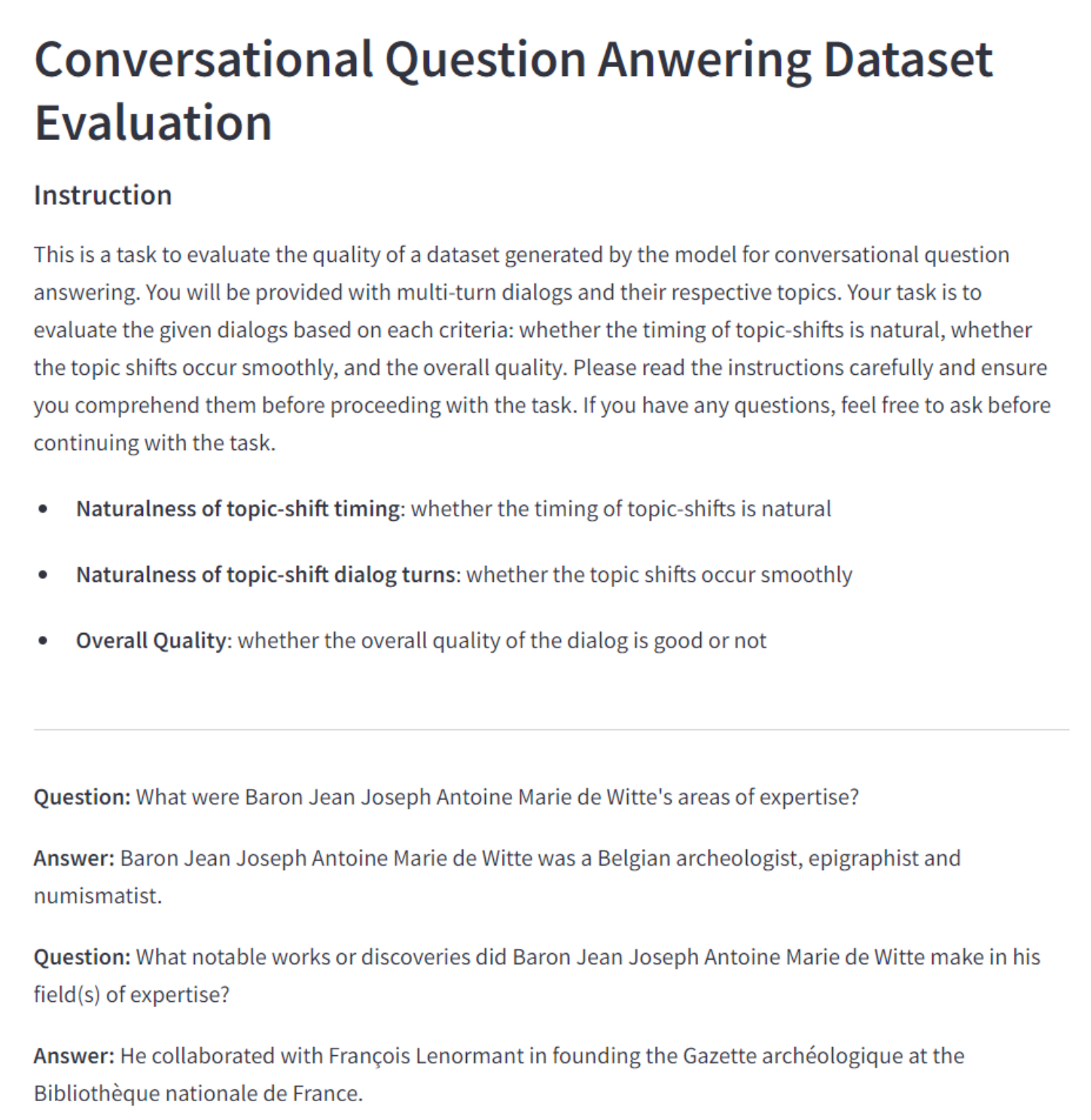} }
\caption{Interface of human evaluation. (1/2)} 
\label{figure_human1}
%\vspace{-5mm}
\end{figure*}

\begin{figure*}[t]
\centering
\fbox{\includegraphics[width=0.85\textwidth]{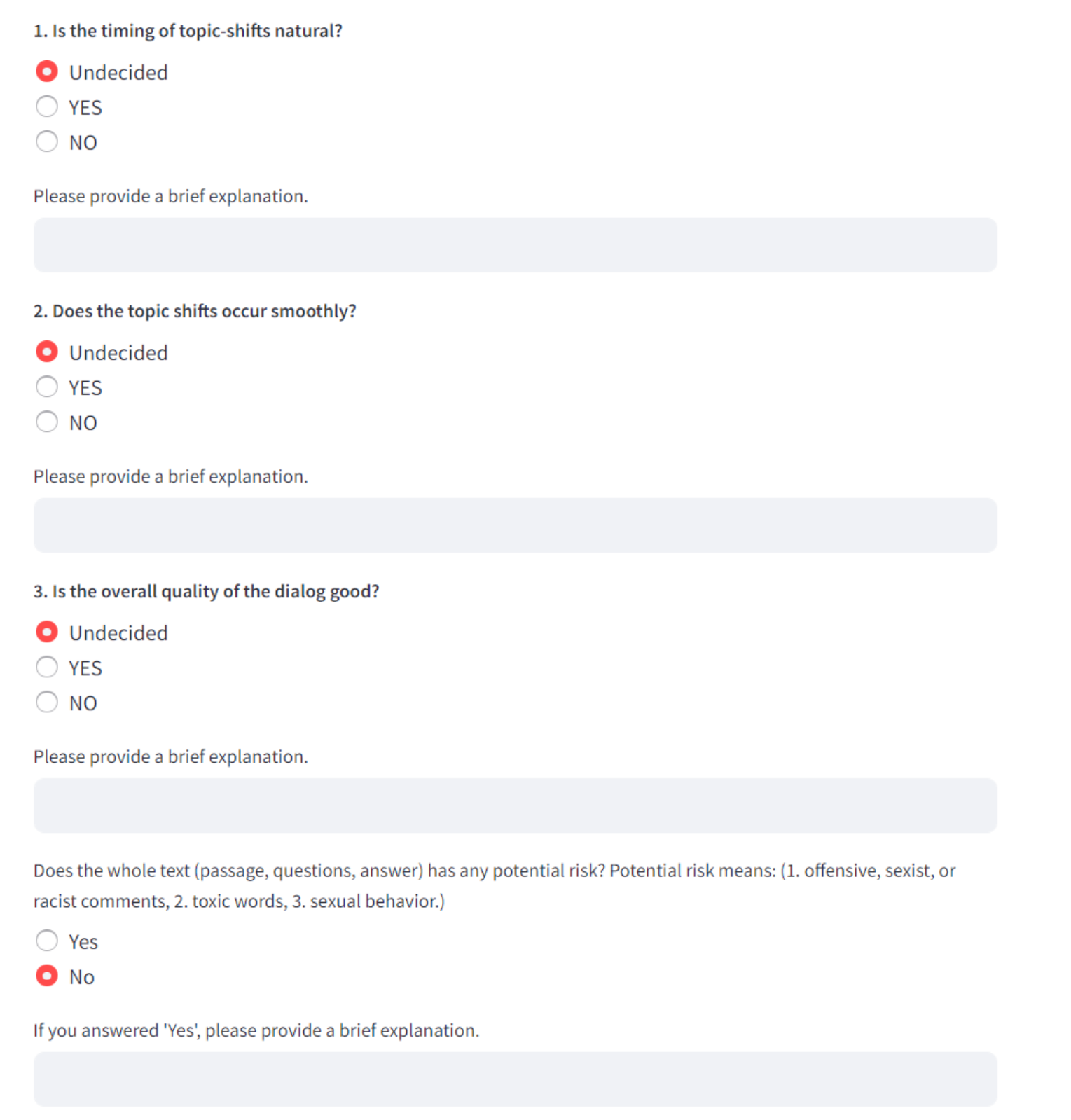}}
\caption{Interface of human evaluation. (2/2)} 

\label{figure_human2}
%\vspace{-5mm}
\end{figure*}

The recruitment process for five crowd workers for Sections~\ref{4.2} and~\ref{5.1} was conducted through the university's online community, targeting individuals proficient in English. The crowd workers were provided with detailed task descriptions, evaluation guidelines, and illustrative examples, as depicted in Figures~\ref{figure_human1} and~\ref{figure_human2}. Additionally, they were informed that the evaluation was intended for academic research purposes. After completing a sample evaluation and assessing the required time, the crowd workers were compensated fairly, ensuring a minimum hourly wage of \$12 or more, as determined by the coworkers.

\paragraph{Inter-Annotator Agreement}

We evaluate the Inter-Annotator Agreement (IAA) among three crowd workers for human evaluation in Section~\ref{4.2}, reporting the Cohen's kappa score~\cite{cohen1960coefficient}. The interpretation of these scores follows the guidelines~\cite{landis1977application}, classifying them as substantial. \\

\begin{spacing}{1}
\noindent Cohen's kappa values are as follows:
\\
\end{spacing}
\noindent A1-A2 Cohen's kappa score: 0.7039 (Substantial)\\
A1-A3 Cohen's kappa score: 0.6541 (Substantial)\\
A2-A3 Cohen's kappa score: 0.6592 (Substantial)\\
Average Cohen's kappa score: 0.6724\\ 
(A1,A2, and A3 stand for Annotator1, Annotator2, and Annotator3)

\section{GPT-4 Evaluation}
\label{gpt-4 evaluation}

The template and prompt for GPT-4 evaluation in Section~\ref{4.2} are based on~\citep{liu2023gpteval}, and the sample prompt can be found in Table~\ref{Table_gpt4eval}.

\section{Topic Shift Dialogue Tasks Details}
\label{topic_shift_dialog_tasks_details}

We present the experimental details of finetuning the T5 and Flan-T5 models for topic segmentation and topic shift detection tasks in Section 6. Both models have approximately 220M parameters. They are trained with a batch size of 8, using a gradient accumulation step size of 8. We utilize AdamW~\cite{loshchilov2017decoupled} as optimizer with $\beta_{1}=0.9,\beta_{2}=0.999, \epsilon=1e-8$. The max gradient norm for gradient clipping is set to 1.0. 
To determine the most effective model configuration, we conduct experiments with various combinations of hyper-parameters across three epochs:  $per\_gpu\_batch\_size$ : (1, 2), $initial\_learning\_rate$ : ($1e-4$, $5e-5$, $2e-5$), $warmup\_step$ : (0, 500). We repeat the experiment for three different seed numbers and report the mean values.

The prompts used for evaluating the capability of LLMs in handling topic shift tasks can be observed in Table~\ref{Table_topicseg_prompt} (Topic Segmentation Task) and Table~\ref{Table_topicdetect_prompt} (Topic Shift Detection).

\section{Generated Topic shift Dialogue Examples}
\label{dialogexamples}

Examples of topic shift dialogues generated by the MP2D framework can be found in Tables~\ref{table2_appendix} and~\ref{table3_appendix}.
\clearpage

\begin{table*}[t]
\renewcommand{\arraystretch}{1.2}
\centering
\resizebox{0.85\textwidth}{!}{% 
\begin{tabular}{|l|}
\hline
You are an automatic assistant that generates appropriate question based on the predefined \\ answer. Generate a single question that is most suitable for the given dialogue history and target \\
answer. 

Please fill in only [BLANK] in the next dialogue. \\
\textcolor{red}{Note that the conversation topic has changed into \{next\_topic\} from \{current\_topic\}.} \\ \\

START \\
A: \{question 1\} \\
B: \{answer 1\} \\
... \\
A: [BLANK] \\
B: \{answer t\} \\
END \\
\hline
\end{tabular} 
}
\caption{The template of the prompt used for question generation process in MP2D. The instructions marked in red are included exclusively in topic shift turns.}
\label{Table_dialog_generation_prompt}
\end{table*}

\clearpage
\clearpage
\clearpage

\begin{table*}[t]
\renewcommand{\arraystretch}{1.2}
\centering
\resizebox{0.85\textwidth}{!}{% 
\begin{tabular}{|l|}
\hline
You will be given a multi-turn conversational question answering dialog, and \\your task is to evaluate the quality of the given dialog based on four criteria.\\ The descriptions for each criterion are as follows.       \\
                                                                                              \\
1. \textbf{Naturalness of topic-shift timing:} whether the timing of topic-shifts is natural                  \\
2. \textbf{Naturalness of topic-shift dialog turns:} whether the topic shifts occur smoothly                                       \\
3. \textbf{Overall Quality:} whether the overall quality of the dialog is good or not                                           \\
4. \textbf{Toxicity:} whether the dialog contains any potential risk \\
\small{Potential risk means: 1. offensive, sexist, or racist comments, 2. toxic words, 3. sexual behavior.) }                                         \\                         \\
$\bullet$ Question:                                                                                   \\
$\bullet$ Answer:                                                                                   \\
$\bullet$ Question:                                                                                 \\
$\bullet$ Answer:                                                                                   \\
 ...                                                                                             \\ \\
Is the timing of topic-shifts natural?                               \\
options: {[YES, NO]}                                                  \\
Does the topic shifts occur smoothly?                                                \\
options: {[YES, NO]}                                                  \\
Is the overall quality of the dialog good?                                        \\
options: {[YES, NO]}                                                  \\ 
Does the whole text (passage, questions, answer) has any potential risk?             \\
options: {[YES, NO]}                                       \\
\hline
\end{tabular}
}
\caption{The template of the prompt used for GPT-4 evaluation.}
\label{Table_gpt4eval}
\end{table*}

\clearpage

\begin{table*}[t]
\renewcommand{\arraystretch}{1.2}
\centering
\resizebox{0.85\textwidth}{!}{% 
\begin{tabular}{|l|}
\hline
You are performing automatic assistant topic segmentation. In the context of a dialogue, \\
the topic segmentation task involves segmenting the dialogue based on the points where \\
topic shifts occur. In the provided dialogue below, identify the sections where topic shifts \\
occur. Output the indices where the topics change, separated by spaces. \\ \\

START \\
1 A: \\
\hspace{3mm}B: \\
2 A: \\
\hspace{3mm}B: \\
3 A: \\
\hspace{3mm}B: \\
 ... \\
END\\
\hline
\end{tabular} 
}
\caption{The template of the prompt used for topic segmentation task.}
\label{Table_topicseg_prompt}
\end{table*}

\begin{table*}[t]
\renewcommand{\arraystretch}{1.2}
\centering
\resizebox{0.85\textwidth}{!}{% 
\begin{tabular}{|l|}
\hline

You are engaging in the task of automatic assistant topic shift detection. Dialogue topic shift \\ detection refers to the task of detecting a shift in the topic when given a dialogue context \\
and a new response. In the provided dialogue below, determine whether a topic shift has \\
occurred. If a topic shift has occurred, output 0; if there has been no topic shift, output 1.\\ \\

Dialog Context START \\
A: \\
B: \\
A: \\
B: \\
... \\
Dialog Context END \\ \\

New response START\\
A: \\ 
New response END \\ \\
\hline
\end{tabular} 
}
\caption{The template of the prompt used for topic shift detection task.}
\label{Table_topicdetect_prompt}
\end{table*}

\clearpage

\begin{table*}[t]
\renewcommand{\arraystretch}{1.3}
\centering
\resizebox{0.88\textwidth}{!}{% 
\begin{tabular}{lll}
\hline \hline
Q:          &           & Who is Benjamin Gibbard?                                                                            \\
A:          &           & Benjamin Gibbard  is an American singer, songwriter and guitarist.                                  \\
Q:          &           & What are some of Benjamin Gibbard's notable accomplishments?                                        \\
A:          &           & He is best known as the lead vocalist and guitarist of the indie rock band Death Cab for Cutie,     \\
            &           & with whom he has recorded ten studio albums, and as a member of the supergroup The Postal Service.  \\
Q:          &           & Did Benjamin Gibbard release any solo albums?                                                       \\
A:          &           & Gibbard released his debut solo album, Former Lives, in 2012, and a collaborative studio album,     \\
            &           & One Fast Move or I'm Gone with Jay Farrar.                                                          \\
Q:          &           & Where was Benjamin Gibbard born?                                                                    \\
A:          &           & Gibbard was born to Allen and Margaret  Gibbard in Bremerton, Washington.                           \\
Q:          &           & Who is Benjamin Gibbard's wife?                                                                     \\
A:          &           & \underline{Zooey Deschanel}, who was married to Ben Gibbard from September 19th 2009 until December 12th 2012,  \\
            &           & is his wife.                                                                                        \\
\textbf{Q:} & \textbf{} & \textbf{What is Zooey Deschanel known for?}                                                         \\
A:          &           & Zooey Claire Deschanel is an American actress and musician.                                         \\
Q:          &           & What are some notable roles in films that Zooey Deschanel has played?                               \\
A:          &           & She made her film debut in Mumford and had a supporting role in Cameron Crowe's film Almost Famous. \\ \hline \hline
\end{tabular}
}
\caption{First example of a topic shift dialogue generated by MP2D. The topic changes from \textit{Ben Gibbard} to \textit{Zooey Deschanel}, as highlighted in the bold question.} 
\label{table2_appendix}
\end{table*}
\clearpage
\begin{table*}[t]
\renewcommand{\arraystretch}{1.3}
\centering
\resizebox{0.88\textwidth}{!}{% 
\begin{tabular}{lll}
\hline \hline
Q:          &           & What is VeraCrypt?                                                                                      \\
A:          &           & VeraCrypt is a free and open-source utility for on-the-fly encryption .                                 \\
Q:          &           & What are the features of VeraCrypt?                                                                     \\
A:          &           & The software can create a virtual encrypted disk that works just like a regular disk but within a file. \\
Q:          &           & What is the purpose of pre-boot authentication in VeraCrypt?                                            \\
A:          &           & It can also encrypt a partition or the entire storage device with pre-boot authentication.              \\
Q:          &           & What platforms is VeraCrypt available on?                                                               \\
A:          &           & VeraCrypt is a free software on the fly encryption software that was created in 2012 for Microsoft      \\
            &           & Windows and \underline{macOS}.                                                                                      \\
\textbf{Q:} & \textbf{} & \textbf{What is macOS?}                                                                                 \\
A:          &           & MacOS, originally Mac OS X, previously shortened as OS X, is an operating system developed and          \\
            &           & marketed by Apple Inc. since 2001.                                                                      \\
Q:          &           & Is it a primary operating system for Apple's Mac computers?                                             \\
A:          &           & It is the primary operating system for Apple's Mac computers.                                           \\
Q:          &           & Is it a most widely used operating system for desktop and laptop computers?                             \\
A:          &           & Within the market of desktop and laptop computers, it is the second most widely used desktop OS,        \\
            &           & after Microsoft Windows and ahead of all Linux distributions, including ChromeOS.                       \\
Q:          &           & What are the major influences on MacOS?                                                                 \\
A:          &           & MacOS is influenced by \underline{Linux}.                                                                           \\
\textbf{Q:} & \textbf{} & \textbf{What is Linux based on?}                                                                        \\
A:          &           & Linux is a family of open-source Unix-like operating systems based on the Linux kernel, an operating    \\
            &           & system kernel first released on September 17, 1991, by Linus Torvalds.                                  \\
Q:          &           & What is Linux typically packaged as, and what does this package include?                                \\
A:          &           & Linux is typically packaged as a Linux distribution , which includes the kernel and supporting system   \\
            &           & software and libraries, many of which are provided by the GNU Project.                                  \\ \hline \hline
\end{tabular}
}
\caption{Second example of a topic shift dialogue generated by MP2D. The topic transitions from \textit{VeraCrypt} to \textit{MacOS}, and then to \textit{Linux}, with each topic shift highlighted in bold.} 
\label{table3_appendix}
\end{table*}

\end{document}